\documentclass[letterpaper]{article} 
\PassOptionsToPackage{table}{xcolor}

\usepackage{aaai2027}  
\usepackage[hyphens]{url}  
\usepackage{graphicx} 
\usepackage{natbib}  
\usepackage{caption} 
\usepackage{multirow}
\usepackage{amsmath}
\usepackage{amssymb}
\usepackage{cleveref}
\usepackage{booktabs}
\usepackage{multirow}
\usepackage{graphicx}

\usepackage{booktabs}
\usepackage{pifont}

\newcommand{\cmark}{\ding{51}}
\newcommand{\xmark}{\ding{55}}
\usepackage{array}
\definecolor{actblue}{RGB}{235,244,252}
\definecolor{dpgreen}{RGB}{237,248,239}
\definecolor{piorange}{RGB}{253,243,230}

\usepackage{algorithm}
\usepackage{algorithmic}

\usepackage{newfloat}
\usepackage{listings}
\DeclareCaptionStyle{ruled}{labelfont=normalfont,labelsep=colon,strut=off} 
\floatstyle{ruled}
\newfloat{listing}{tb}{lst}{}
\floatname{listing}{Listing}

\usepackage{booktabs}

\title{CAAT: Contact-Aware Attention Scaling and Tactile Masking for \\Data-Efficient Contact-Rich Manipulation}

\author{\normalfont
    Jiaming Jiang\textsuperscript{\rm 1,3$\star$},
    Yuzhe Huang\textsuperscript{\rm 2,3$\star$}, 
    Hao Liang\textsuperscript{\rm 3}, 
    Pei Lin\textsuperscript{\rm 1,3}, 
    Shengcheng Luo\textsuperscript{\rm 1,3}, 
    \\
    Fanrong Dong\textsuperscript{\rm 4}, 
    Jiaping Wu\textsuperscript{\rm 5}, 
    Chenxi Xiao\textsuperscript{\rm 1$\dagger$},    
    Wanlin Li\textsuperscript{\rm 3$\dagger$},
    Ziyuan Jiao\textsuperscript{\rm 2,3$\dagger$}
    \\
    \textsuperscript{\rm 1}ShanghaiTech University \quad
    \textsuperscript{\rm 2}Beihang University \quad
    \\
    \textsuperscript{\rm 3}Beijing Institute for General Artificial Intelligence \quad
    \textsuperscript{\rm 4}Zhejiang University \quad
    \textsuperscript{\rm 5}BUPT \quad
    \\
    \textsuperscript{$\star$}Equal contributors \quad 
    \textsuperscript{$\dagger$}Corresponding authors \quad 
}
\affiliations{}

\makeatletter \newcommand{\CAATteaser}{%
\par
\begingroup 
\centering \includegraphics[width=0.98\textwidth]{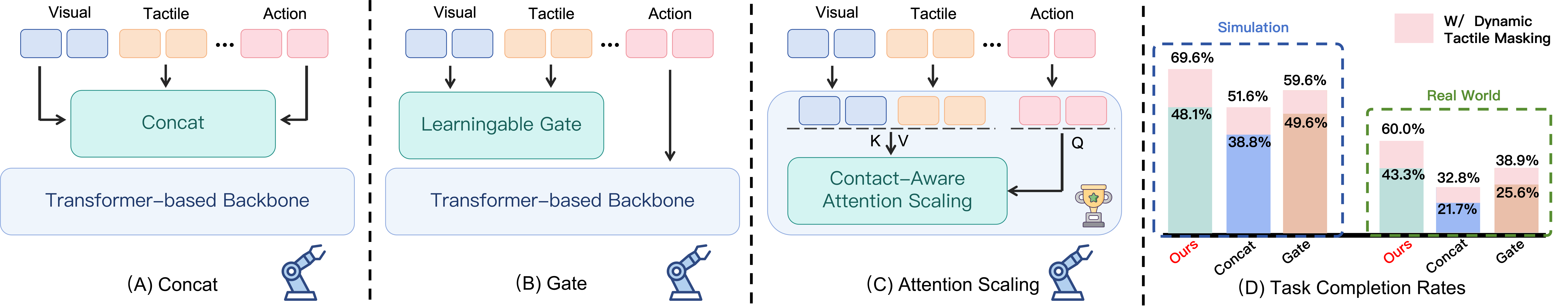} \par \captionof{figure}{ \textbf{Overview of visuo-tactile fusion strategies and their performance comparisons:}  (A) Direct token concatenation and (B) learnable gating lack explicit priors for switching between contact and non-contact phases. (C) CAAT introduces contact-aware attention scaling to emphasize vision tokens before contact and tactile tokens during contact. (D) With dynamic tactile masking applied to all methods, CAAT consistently outperforms concatenation and gated fusion on both simulation and real-world manipulation tasks.} \label{fig:teaser} 
\endgroup 
\par
\medskip }
\g@addto@macro\@maketitle{\CAATteaser} 
\makeatother

\begin{document}
\maketitle

\begin{abstract}
In contact-rich manipulation, visual observations primarily guide motion in free space, whereas tactile observations become particularly informative during contact. However, standard Transformer-based visuo-tactile policies typically rely on either token concatenation or learnable gating. These approaches lack explicit contact-aware priors, making it difficult to efficiently learn effective cross-modal representations from demonstrations. To address this limitation, we propose CAAT, a lightweight contact-aware framework that explicitly incorporates contact priors through attention scaling and dynamic tactile masking. Specifically, CAAT emphasizes visual information before contact and tactile information during contact. It also suppresses static background tokens by comparing the current tactile observation with a non-contact reference. CAAT can be integrated into commonly used Transformer-based policies without modifying their action decoders. In simulation, integrating CAAT with ACT improves the average success rate by 18.0 percentage points over direct visuo-tactile fusion and by 10.0 percentage points over gated fusion. In real-world experiments using a visuo-tactile UMI platform, CAAT achieves an average success rate of $60.0\%$ across ACT, Diffusion Policy, and $\pi_0$, outperforming the strongest baseline by an average of 21.1 percentage points. These results demonstrate that explicit contact priors and dynamic tactile masking are effective in improving visuo-tactile policy learning and task performance of diverse policy architectures. 
\url{https://mrjiangjm.github.io/caat/}

\end{abstract}


\section{Introduction}

Contact-rich manipulation has long posed challenges for learning effective visuo-tactile policies~\citep{chen2022visuo,khanh2025manifeel,bi2026vla}. During much of a manipulation trajectory, the robot operates in free space or performs coarse alignment, for which visual observations provide sufficient global information about object locations~\citep{team2024octo} and scene geometry~\citep{zitkovich2023rt}. By contrast, tactile sensing is most informative during contact phases~\citep{zhang2026tacvla,yuan2026ftp}, when task success critically depends on local physical interactions, such as applying appropriate contact forces to securely grasp objects~\citep{huang2026tactile,she2021cable,huang2026ht}, preventing incipient slip~\citep{james2020slip}, and regulating forces during pulling or insertion~\citep{huang2026tactile1,hao2025tla}. These observations suggest that, although tactile sensing is indispensable for dexterous manipulation~\citep{heng2025vitacformer}, its utility is inherently stage-dependent~\citep{xue2025reactive,li2020review}.

Existing visuo-tactile policies typically lack such a contact-aware prior~\citep{li2020review}. In Transformer-based policies, a common approach is to encode tactile observations as additional tokens and directly concatenate them with visual and proprioceptive tokens, allowing self-attention to learn cross-modal fusion end-to-end~\citep{zhao2023learning,chi2025diffusion,black2024pi_0,jiang2025gelfusion}. Although highly expressive, as it preserves information from all modalities, this design requires the model to infer from demonstrations both \emph{when} tactile information is relevant and \emph{how} it should be integrated with other sensory inputs. This poses a challenge to the learning process of contact-rich manipulation, where contact events constitute only a small fraction of each trajectory and thus informative tactile signals can easily be overwhelmed by abundant non-contact observations. Recent gated or adaptive fusion methods alleviate this issue by learning modality-dependent weighting schemes~\citep{zang2026tacforesight,li2025adaptive}. Nevertheless, these gating mechanisms must still be learned from the same demonstrations, in which informative contact events are sparse. Consequently, such methods often require large-scale datasets to achieve robust performance~\citep{lou2026dream} and can degrade performance in data-scarce settings.

In this work, our key insight is that contact and non-contact phases can be distinguished relatively easily, allowing modality allocation to be explicitly conditioned on contact state rather than inferred from demonstrations~\citep{zang2026tacforesight,li2025adaptive}. Based on the sensorimotor prior that vision is generally more informative before contact~\citep{lei2026learning}, whereas tactile sensing becomes more informative during contact~\citep{li2026vla}, we propose \textbf{CAAT}, a \textbf{C}ontact-\textbf{A}ware \textbf{A}ttention Scaling and \textbf{T}actile Masking framework. CAAT imposes this prior at the attention readout while preserving expressive cross-modal representation learning. CAAT consists of two components. \textbf{First},
\textbf{Contact-Aware Attention Scaling} constructs modality-specific
visual, tactile, and proprioceptive action-query readouts by separately
attending to the corresponding contextualized token subsets after joint
self-attention. It then aggregates these readouts using
contact-conditioned scaling, emphasizing vision for localization and
approach before contact and tactile information for local interaction
control during contact. Using simple empirical scaling factors reduces the burden of learning modality-allocation rules from scarce demonstrations while introducing negligible computational overhead. \textbf{Second}, \textbf{Dynamic Tactile Masking (DTM)} suppresses unchanged background tokens and retains contact-relevant regions~\citep{li2025controlvla}, enabling the Transformer to focus on contact-induced tactile changes.

Moreover, CAAT is naturally modular and can be integrated into existing Transformer-based policies without modifying their action decoders. We instantiate CAAT on Action Chunking with Transformers (ACT)~\citep{zhao2023learning}, Diffusion Policy (DP)~\citep{chi2025diffusion}, and $\pi_0$~\citep{black2024pi_0}. In simulation, we evaluate CAAT with ACT on the UniVTAC benchmark~\citep{chen2026univtac}, where it improves the average success rate by 18 percentage points over standard visuo-tactile fusion and by 10 percentage points over gated fusion. We further evaluate CAAT with ACT, DP, and $\pi_0$ in real-world contact-rich manipulation using a dexterous two-finger tactile UMI device, where CAAT outperforms the strongest corresponding baseline by 21.1 percentage points on average. These results demonstrate that CAAT provides effective contact-stage priors for utilizing tactile information in contact-rich manipulation while remaining compatible with different policy architectures.

The main contributions of this work are as follows:
\begin{itemize}
    \item We identify the phase-dependent sparsity of tactile feedback as a key challenge in contact-rich manipulation, motivating the need for contact-aware visuo-tactile fusion.

    \item We propose \textbf{CAAT}, a plug-and-play framework that injects contact-stage priors into Transformer-based policies through \textbf{Contact-Aware Attention Scaling} and \textbf{Dynamic Tactile Masking}.

    \item We develop an isomorphic two-finger visuo-tactile UMI device that enables efficient data collection of contact-rich human demonstrations and robot policy deployment.
    
    \item We demonstrate the effectiveness of CAAT with ACT in simulation and further validate its consistent improvements across ACT, DP, and $\pi_0$ in real-world contact-rich manipulation.
\end{itemize}

\section{Related Work}

\subsection{Tactile Sensing for Manipulation}

Tactile sensing provides physical feedback that is difficult to infer from vision alone, particularly in contact-rich manipulation. Prior work has leveraged tactile observations for grasp stabilization and slip detection~\citep{lin2025pp,james2020slip}, force-sensitive control~\citep{yuan2017gelsight}, pose estimation~\citep{wang2025tactape,huang2026tactile1}, and dexterous in-hand manipulation~\citep{lin2026dexmove,luo2026blind}. These studies demonstrate that tactile feedback is especially valuable at the contact interface, where local interaction states may be difficult to observe visually because of occlusion by the hand or manipulated object.

Despite this progress, how tactile information should be integrated into manipulation policies remains an open question~\citep{luo2026rgb,zhang2026vtla}. This challenge is particularly important when contact-rich demonstrations are scarce, as appropriate inductive biases become critical for effective learning in limited-data regimes. Our work addresses this gap by introducing architectural priors tailored to contact-aware visuo-tactile fusion.

\begin{figure*}[t]
    \centering
    \includegraphics[width=\linewidth]{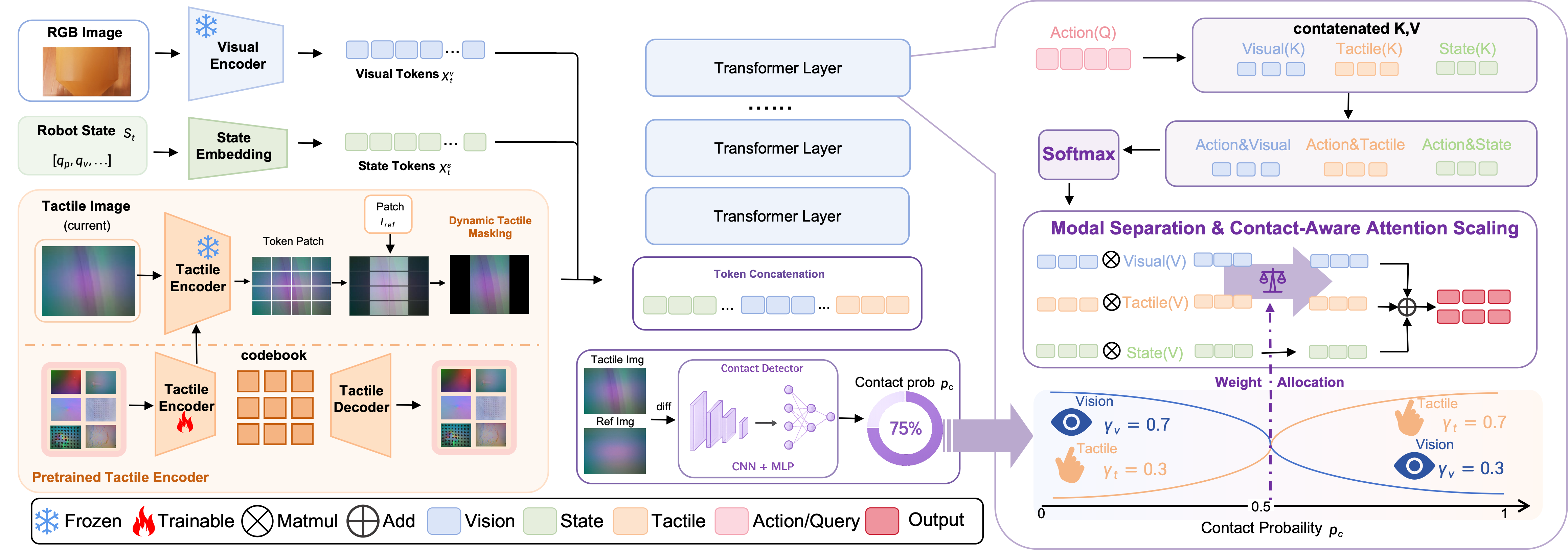}
    \caption{Overview of CAAT. CAAT introduces contact-aware priors into Transformer-based visuo-tactile action prediction. Visual observations and robot states are encoded as modality-specific tokens, while tactile images are converted into patch tokens using a pretrained tactile encoder. CAAT first applies \emph{Dynamic Tactile Masking}, which compares the current tactile tokens with a non-contact reference and suppresses unchanged background regions while retaining tokens that capture contact-induced deformations. The visual, masked tactile, and state tokens are then concatenated and processed by Transformer layers. For action decoding, CAAT computes modality-specific attention readouts and scales the visual and tactile contributions according to the estimated contact state: visual features are emphasized before contact, whereas tactile features are emphasized during contact. The resulting multimodal representation is passed to the action decoder to predict a sequence of future actions.}
    \label{fig:pipeline}
\end{figure*}

\subsection{Visuo-Tactile Fusion in Transformer Policies}

Imitation-learning methods commonly employ Transformer-based architectures to predict actions from multimodal observations~\citep{bi2026vla,cheng2025omnivtla,huang2026tactile,zhang2026tacvla}. These policies use attention mechanisms to integrate visual, tactile, and proprioceptive tokens and predict temporally coherent action sequences~\citep{luo2026rgb,zhao2024transferable}. However, designing effective multimodal fusion strategies remains an open challenge. Recent work has explored adaptive approaches, including attention-based weighting, gating networks~\citep{nagrani2021attention}, mixture-of-experts modules~\citep{cao2023multi}, and confidence-based fusion~\citep{han2022trusted}, to explicitly regulate modality contributions~\citep{ruan2026retac,lou2026dream,zang2026tacforesight}. In contact-rich settings, however, learning reliable modality weights from limited demonstrations remains difficult because informative tactile signals are temporally sparse. CAAT addresses this limitation through contact-aware attention scaling, which uses the estimated interaction phase as a structural prior for data-efficient visuo-tactile fusion.

\section{Method}

\subsection{Problem Formulation and Overview}

\subsubsection{Problem Formulation} We formulate visuo-tactile manipulation as an action sequence prediction problem. At each timestep $t$, the policy observes a history of visual inputs $I_{t-k:t}$, tactile inputs $T_{t-k:t}$, and additional conditioning variables $\Upsilon$, such as proprioception or task information. The objective is to predict a horizon of future actions $a_{t:t+H}$ that complete the manipulation task:
\begin{equation}
    a_{t:t+H} = \pi_\theta \left( I_{t-k:t}, T_{t-k:t}, \Upsilon \right),
\end{equation}
where $k$ denotes the observation history length and $H$ denotes the action prediction horizon.

Standard end-to-end visuo-tactile policies must learn the allocation between visual and tactile modalities directly from demonstrations. This can be data-inefficient, especially when contact occurs only during a small fraction of the trajectory or when the amount of training data is limited. To address this challenge, we introduce \textbf{CAAT}, a lightweight framework that injects contact-aware priors into Transformer-based action prediction models to improve action prediction quality.

\noindent \textbf{Overview of Method.}
The overall pipeline of our proposed approach is shown in
Fig.~\ref{fig:pipeline}. A pretrained tactile encoder first converts
raw tactile observations into spatial token representations.
CAAT then introduces two lightweight components on top of these
representations. First, Dynamic Tactile Masking suppresses unchanged
non-contact background tokens by comparing the current tactile tokens
with a non-contact reference, allowing the policy to focus on
contact-induced changes (Section~3.2). Second, Contact-Aware
Attention Scaling modulates the modality-specific Transformer
readouts according to the estimated contact phase, emphasizing
visual information before contact and tactile information during
contact (Section~3.3). Both components can be integrated into
different Transformer-based policy backbones without modifying
their action decoders.


\subsection{Tactile Preprocessing}
\subsubsection{Tactile Token Representation}\label{method:encoder}

Our approach relies on a tactile encoder that converts raw tactile observations into token representations. Given a tactile observation $T_t$, the tactile encoder $E_\tau$ maps the tactile image into a sequence of patch tokens:
\begin{equation}
    X_t^\tau = E_\tau(T_t), 
    \quad 
    X_t^\tau = [x_{t,1}^\tau,\dots,x_{t,N}^\tau] \in \mathbb{R}^{N \times d},
\end{equation}
where $N$ is the number of tactile patches and $d$ is the token dimension.
In our implementation, $E_\tau$ is a pre-trained tactile ViT encoder. More details of $E_\tau$ are provided in Appendix.

\subsubsection{Dynamic Tactile Masking}\label{method:mask}

Although a pretrained tactile encoder provides effective representations, directly encoding the full tactile observation may include static non-contact background information. To focus the policy on task-relevant deformations caused by physical interaction, we introduce Dynamic Tactile Masking, which masks out non-contact regions in ViT patches. This operation injects a spatial prior: informative tactile evidence is concentrated in regions that change relative to the non-contact state, rather than being distributed across the entire tactile image. By suppressing static background tokens, Dynamic Tactile Masking reduces the burden on the policy to distinguish contact signals from sensor appearance, thereby improving sample efficiency in low-data settings.

Specifically, we first define a reference tactile observation $T_0$ from the initial non-contact phase. Both the current tactile observation and the reference observation are encoded by the same tactile encoder:
\begin{equation}
    X_t^\tau = E_\tau(T_t), 
    \quad 
    X_0^\tau = E_\tau(T_0).
\end{equation}

For each tactile patch token, we compute its cosine similarity to the corresponding reference token:
\begin{equation}
    c_{t,n} =
    \frac{x_{t,n}^\tau \cdot x_{0,n}^\tau}
    {\|x_{t,n}^\tau\| \, \|x_{0,n}^\tau\|},
    \quad n=1,\dots,N.
\end{equation}

The similarity score is then converted into a binary mask:
\begin{equation}
M_{t,n} =
\begin{cases}
0, & c_{t,n} \geq \rho, \\
1, & c_{t,n} < \rho,
\end{cases}
\end{equation}
where $\rho$ is a similarity threshold. Tokens with high similarity to the reference are treated as unchanged background, while tokens with low similarity are treated as contact-induced tactile changes. The masked tactile token is computed as:
\begin{equation}
    \tilde{x}_{t,n}^\tau = M_{t,n} x_{t,n}^\tau.
\end{equation}
Thus, high-similarity background tokens are suppressed, while tokens likely to contain contact deformation are retained. The resulting masked tactile sequence is
\begin{equation}
    \tilde{X}_t^\tau =
    [\tilde{x}_{t,1}^\tau,\dots,\tilde{x}_{t,N}^\tau],
\end{equation}
which is used as the tactile input to the Transformer policy.

\subsection{Contact-Aware Attention Scaling}\label{method:guide}
We next adaptively regulate the contributions of visual and tactile information before and during contact. To this end, we propose Contact-Aware Attention Scaling, which controls when and to what extent tactile observations contribute to the policy readout during manipulation.

For clarity, we present CAAT using a generic token-based formulation.
The exact representation and injection pathway of proprioceptive,
latent, and action-conditioning features depend on the underlying
policy architecture.

Given visual tokens $X_t^v$, masked tactile tokens $\tilde{X}_t^\tau$, proprioceptive tokens $X_t^\Upsilon$ and action tokens ${X}_t^a$, we first concatenate all tokens and process them with Transformer self-attention:
\begin{equation}
    X_t = [X_t^v, \tilde{X}_t^\tau, X_t^{\Upsilon},{X}_t^a],
    \quad
    H_t = \mathrm{SelfAttn}(X_t).
\end{equation}
The encoded tokens are then split according to their original modalities:
\begin{equation}
    H_t = [H_t^v, H_t^\tau, H_t^{\Upsilon},H_t^a],
\end{equation}
where $H_t^v$, $H_t^\tau$, $H_t^\Upsilon$, and $H_t^a$
denote the contextualized visual, tactile, proprioceptive, and
action tokens, respectively. This step enables the model to learn contextualized cross-modal representations through self-attention.

We then extract modality-specific cross-attention readouts using an action query $Q_t^{a}=H_t^a W_Q$. For each modality $m \in \{v,\tau,\Upsilon\}$, the corresponding tokens are projected into keys and values:
\begin{equation}
    K_t^m = H_t^m W_K,
    \quad
    V_t^m = H_t^m W_V.
\end{equation}
The action query attends to each modality separately:
\begin{equation}
    w_t^m =
    \mathrm{softmax}
    \left(
    \frac{Q_t^a (K_t^m)^\top}{\sqrt{d_h}}
    \right),
\end{equation}
producing a modality-specific readout:
\begin{equation}
    r_t^m = w_t^m V_t^m,
    \quad
    m \in \{v,\tau,\Upsilon\}.
\end{equation}

Let $z_t$ denote the estimated contact state, where $z_t=0$ indicates the non-contact phase and $z_t=1$ indicates the contact phase. CAAT combines the modality-specific readouts using contact-aware scaling:
\begin{equation}
    \hat{r}_t =
    \gamma_v(z_t) r_t^v
    +
    \gamma_\tau(z_t) r_t^\tau
    +
    r_t^\Upsilon.
    \label{eq:CAAT:scaling}
\end{equation}
The scaling factors $\gamma(\cdot)$ are user-specified functions that encode the desired phase-dependent modality preference. The choice of these functions are detailed in the experiment settings. Generally, these functions satisfy the following rules, indicating that the final representation emphasizes visual information before contact and tactile information during contact:
\begin{equation}
    \gamma_v(0) > \gamma_\tau(0),
    \quad
    \gamma_\tau(1) > \gamma_v(1).
\end{equation}

The contact phase is estimated by a lightweight CNN--MLP binary
classifier operating on the difference between the current tactile
observation and a non-contact reference observation. Specifically,
given the tactile observation \(T_t\) at time step \(t\) and the
reference observation \(T_{\mathrm{ref}}\), we compute
\begin{equation}
    D_t = \left|T_t - T_{\mathrm{ref}}\right|,
\end{equation}
where \(T_{\mathrm{ref}}\) is recorded before interaction begins.
The contact detector takes \(D_t\) as input and predicts a contact
probability \(p_t \in [0,1]\). It is trained with binary cross-entropy
supervision using ground-truth contact labels
\(y_t \in \{0,1\}\):
\begin{equation}
    \mathcal{L}_{\mathrm{contact}}
    =
    -y_t \log p_t
    -(1-y_t)\log(1-p_t).
\end{equation}
During train and inference, the robot is considered to be in contact when
\(p_t > 0.5\), i.e.,
\begin{equation}
    z_t = \mathbb{I}\left(p_t > 0.5\right).
\end{equation}
The resulting binary contact state \(z_t\) is used to switch the
visual and tactile scaling weights in the proposed contact-aware
attention readout.


Finally, the scaled representation $\hat{r}_t$ is passed to the policy action decoder to predict the future action sequence:
\begin{equation}
    a_{t:t+H} = D_\theta(\hat{r}_t),
\end{equation}
where $D_\theta$ denotes the action decoder of the chosen Transformer-based policy backbone.

\begin{table*}[t]
\centering

\begin{tabular}{l|ccccc|c}
\toprule[1.2pt]
\textbf{Method}
& \textbf{Lift Bottle}
& \textbf{Pull Out Key}
& \textbf{Lift Can}
& \textbf{Put Bottle in Shelf}
& \textbf{Insert Tube}
& \textbf{Average} \\
\midrule[0.6pt]

ACT+DTM
& 20\% & 70\% & 59\% & 14\% & 95\% & 51.6\% \\

Gate ACT+DTM
& 65\% & 75\% & 57\% & 11\% & 90\% & 59.6\% \\

Ours (Binary)
& 56\% & 46\% & 50\% & 16\% & 91\% & 51.8\% \\

Ours (Learnable)
& 56\% & 42\% & 51\% & 10\% & 94\% & 50.6\% \\

\textbf{Ours (Numerical)}
& \textbf{71\%}
& \textbf{80\%}
& \textbf{63\%}
& \textbf{38\%}
& \textbf{96\%}
& \textbf{69.6\%} \\

\bottomrule[1.2pt]
\end{tabular}
\caption{Comparison of visuo-tactile fusion methods using ACT as the policy backbone on the UniVTAC manipulation tasks.}
\label{tab:sim_results}
\end{table*}














\begin{table*}[h]
\centering

\setlength{\tabcolsep}{7.65pt}

\begin{tabular}{
l
>{\columncolor{actblue}}c
>{\columncolor{actblue}}c
>{\columncolor{actblue}}c
>{\columncolor{dpgreen}}c
>{\columncolor{dpgreen}}c
>{\columncolor{dpgreen}}c
>{\columncolor{piorange}}c
>{\columncolor{piorange}}c
>{\columncolor{piorange}}c
}
\toprule

\multirow{2}{*}{\textbf{Task}}
& \multicolumn{3}{c}{\cellcolor{actblue}\textbf{ACT Backbone}}
& \multicolumn{3}{c}{\cellcolor{dpgreen}\textbf{Diffusion Backbone}}
& \multicolumn{3}{c}{\cellcolor{piorange}\textbf{$\pi_0$ Backbone}} \\

\cmidrule(lr){2-4}
\cmidrule(lr){5-7}
\cmidrule(lr){8-10}

& \textbf{Concat}
& \textbf{Gate}
& \textbf{Ours}
& \textbf{Concat}
& \textbf{Gate}
& \textbf{Ours}
& \textbf{Concat}
& \textbf{Gate}
& \textbf{Ours} \\

\midrule

Lift Bottle
& 15\% & 20\% & \textbf{40\%}
& 25\% & 35\% & \textbf{35\%}
& 20\% & 15\% & \textbf{50\%} \\

Open Box
& 10\% & 10\% & \textbf{50\%}
& 20\% & 25\% & \textbf{60\%}
& 25\% & 35\% & \textbf{60\%} \\

Powerbank Extraction
& 50\% & 60\% & \textbf{75\%}
& 65\% & 65\% & \textbf{80\%}
& 65\% & 85\% & \textbf{90\%} \\

\midrule

\textbf{Average}
& 25.0\% & 30.0\% & \textbf{55.0\%}
& 36.7\% & 41.7\% & \textbf{58.3\%}
& 36.7\% & 45.0\% & \textbf{66.7\%} \\

\bottomrule
\end{tabular}
\caption{Success rates (\%) of different visuo-tactile fusion methods across three backbones on real-world manipulation tasks.}
\label{tab:real_results}
\end{table*}

\section{Experiments}
\subsection{Experimental Setup}
\label{sec:exp_setup}
\begin{figure}[h!]
    \centering
    \includegraphics[width=\linewidth]{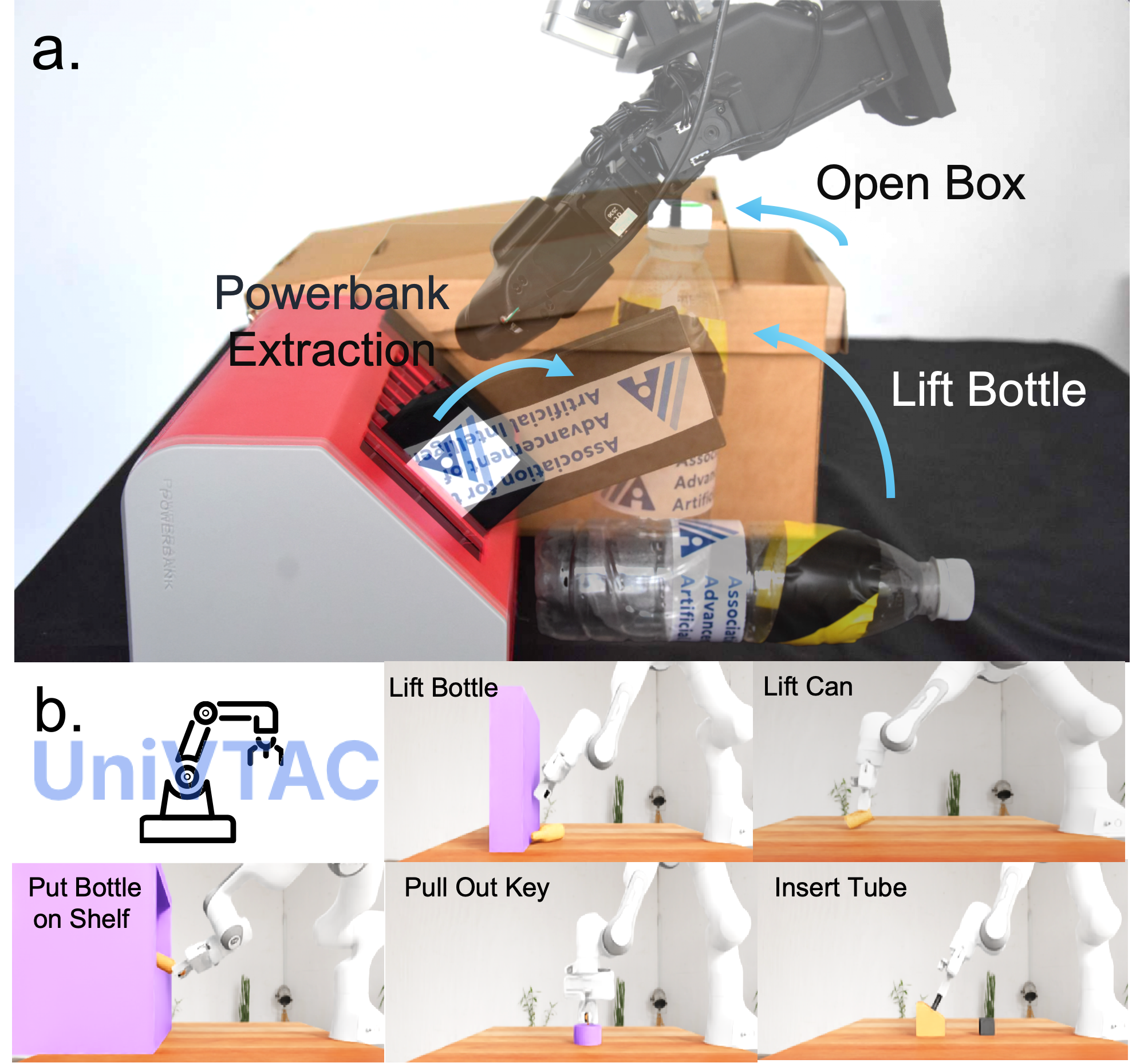}
    \caption{Evaluation settings and tasks. (a) Our proposed approach is evaluated on three real-world contact-rich manipulation tasks: Lift Bottle, Open Box, and Powerbank Extraction. (b) UniVTAC simulation environments used for evaluation.}
    \label{fig:setup}
\end{figure}

To evaluate the effectiveness of CAAT, we conduct experiments in both simulation and real-world settings (Fig.~\ref{fig:setup}). In simulation, we use the UniVTAC benchmark~\citep{chen2026univtac} to compare three visuo-tactile fusion strategies: (1) direct concatenation of visual and tactile features~\citep{chen2026univtac}, (2) fusion through a learnable modality gate~\citep{zang2026tacforesight}, and (3) our proposed CAAT framework. We further evaluate three variants of the scaling function $\gamma$: binary scaling $(0,1)$, learnable scaling, and fixed numerical scaling $(0.3,0.7)$ (see Eq.~\ref{eq:CAAT:scaling}). Implementation details of the learnable-gating baseline and the
learnable-scaling variant are provided in
Appendix.

For real-world evaluation, we develop a UMI-inspired, human-operated two-finger tactile gripper as our experimental platform; hardware details are provided in Appendix. Inspired by human pinch grasping, the gripper enables intuitive handheld collection of contact-rich demonstrations. The same gripper embodiment and tactile-sensing configuration are used during robot deployment, reducing the embodiment gap between demonstration collection and policy execution. Each fingertip provides high-resolution tactile observations during physical interaction. To assess the generalizability of CAAT across policy architectures, we evaluate three representative backbones: ACT, Diffusion Policy (DP), and $\pi_0$.

All models are trained using 150 demonstrations per task in both simulation and real-world experiments. Dynamic Tactile Masking is applied consistently across all fusion methods to ensure a fair comparison. Additional implementation and training details are provided in Appendix.

\begin{figure}[t!]
    \centering
    \includegraphics[width=\linewidth]{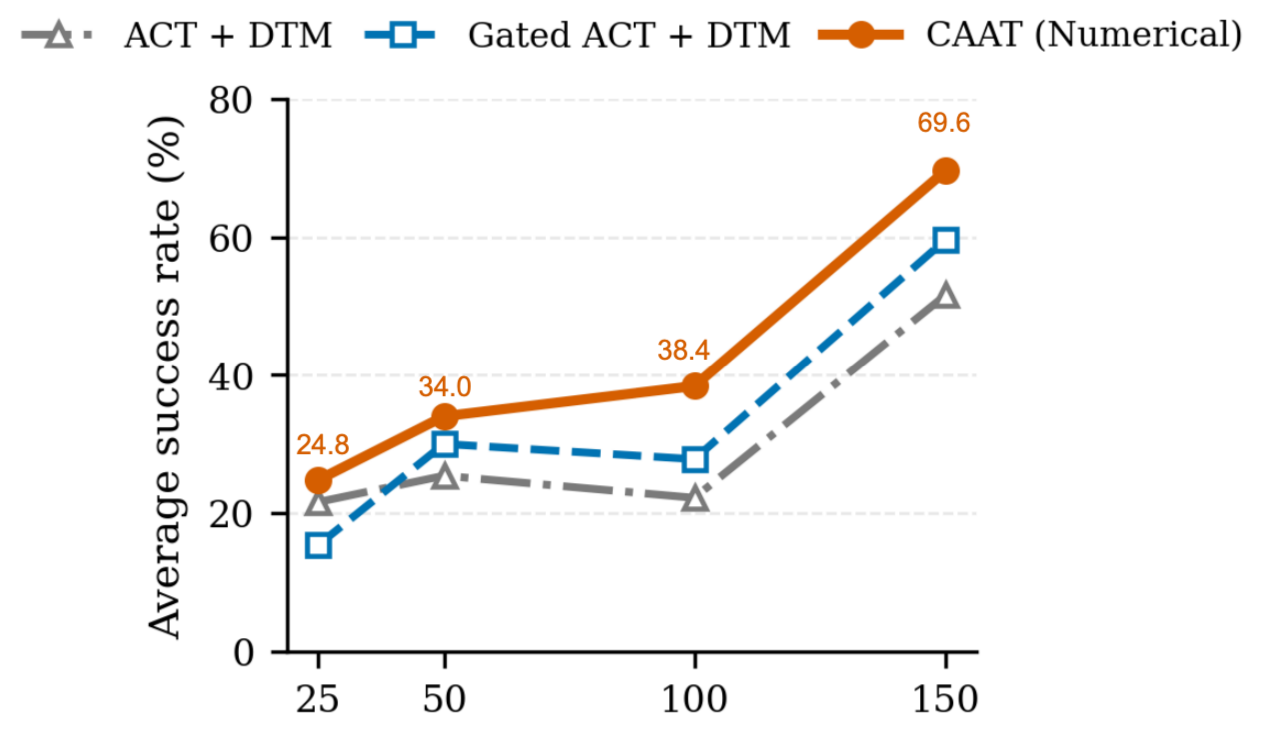}
    \caption{
    Data efficiency on the UniVTAC benchmark. The curves report the
    average success rate across five simulation tasks under different
    numbers of training demonstrations. All methods use ACT as the policy
    backbone and apply Dynamic Tactile Masking. CAAT uses the fixed
    numerical scaling strategy.
    }
    \label{fig:data_efficiency}
\end{figure}

\subsection{Simulation Results on UniVTAC}
\label{sec:sim_results}

\paragraph{Overall performance.}
We first evaluate CAAT on the UniVTAC simulation
benchmark~\citep{chen2026univtac}. \Cref{tab:sim_results} reports the
success rates of different visuo-tactile fusion methods across five
manipulation tasks, all using ACT as the policy backbone. Dynamic
Tactile Masking (DTM) is applied to all methods to ensure a controlled
comparison. CAAT with fixed numerical scaling achieves the best
performance on all five tasks, attaining an average success rate of
$69.6\%$. It outperforms the direct-concatenation baseline
(ACT + DTM) and the learnable-gating baseline (Gated ACT + DTM) by
$18.0$ and $10.0$ percentage points, respectively. The improvements
over ACT + DTM are particularly pronounced on \emph{Lift Bottle} and
\emph{Put Bottle in Shelf}, reaching $51$ and $24$ percentage points,
respectively. Compared with Gated ACT + DTM, CAAT achieves its largest
gain of $27$ percentage points on \emph{Put Bottle in Shelf}. These
results demonstrate the effectiveness of contact-aware attention
scaling when tactile masking is controlled across methods.

\paragraph{Comparison of scaling strategies.}
We further compare three implementations of the scaling function
$\gamma$: binary, learnable, and fixed numerical scaling. Fixed
numerical scaling performs best, achieving an average success rate of
$69.6\%$, whereas the binary and learnable variants achieve $51.8\%$
and $50.6\%$, respectively. This result suggests that fixed numerical
scaling provides a balanced phase-dependent preference without
completely suppressing either modality or requiring the weighting rule
to be learned from limited demonstrations. We therefore adopt fixed
numerical scaling in the subsequent real-world experiments.

\paragraph{Data efficiency.}
We evaluate all methods using 25, 50, 100, and 150 demonstrations under
the same training settings. As shown in \Cref{fig:data_efficiency},
CAAT consistently achieves the highest average success rate across all
demonstration budgets and improves steadily as more data become
available. Under the reduced-data settings of 25, 50, and 100
demonstrations, CAAT achieves $24.8\%$, $34.0\%$, and $38.4\%$,
respectively, outperforming both baselines at every data scale. In
particular, with 100 demonstrations, CAAT exceeds ACT + DTM and Gated
ACT + DTM by $16.2$ and $10.6$ percentage points, respectively. These
results indicate that the contact-aware prior enables more reliable
policy learning when demonstrations are limited. Complete results are
provided in Appendix.

\subsection{Real-World Results}
\label{sec:real_results}

\Cref{tab:real_results} reports the real-world success rates of three policy backbones across three contact-rich manipulation tasks. CAAT achieves the best or tied-best performance in all backbone--task combinations, consistently outperforming direct-concatenation and learnable-gating baselines. Compared with direct concatenation, CAAT increases the average success rate across tasks from $25.0\%$ to $55.0\%$ for ACT, from $36.7\%$ to $58.3\%$ for Diffusion Policy, and from $36.7\%$ to $66.7\%$ for $\pi_0$.

The improvements are particularly pronounced on \emph{Open Box}, where CAAT achieves success rates of $50\%$--$60\%$, compared with $10\%$--$35\%$ for the baselines. CAAT also yields consistent gains on \emph{Powerbank Extraction}, achieving a success rate of up to $90\%$ with $\pi_0$. These results demonstrate that CAAT generalizes across diverse policy architectures and effectively leverages tactile feedback for real-world contact-rich manipulation.

\begin{figure*}[t]
\centering
\includegraphics[width=\textwidth]{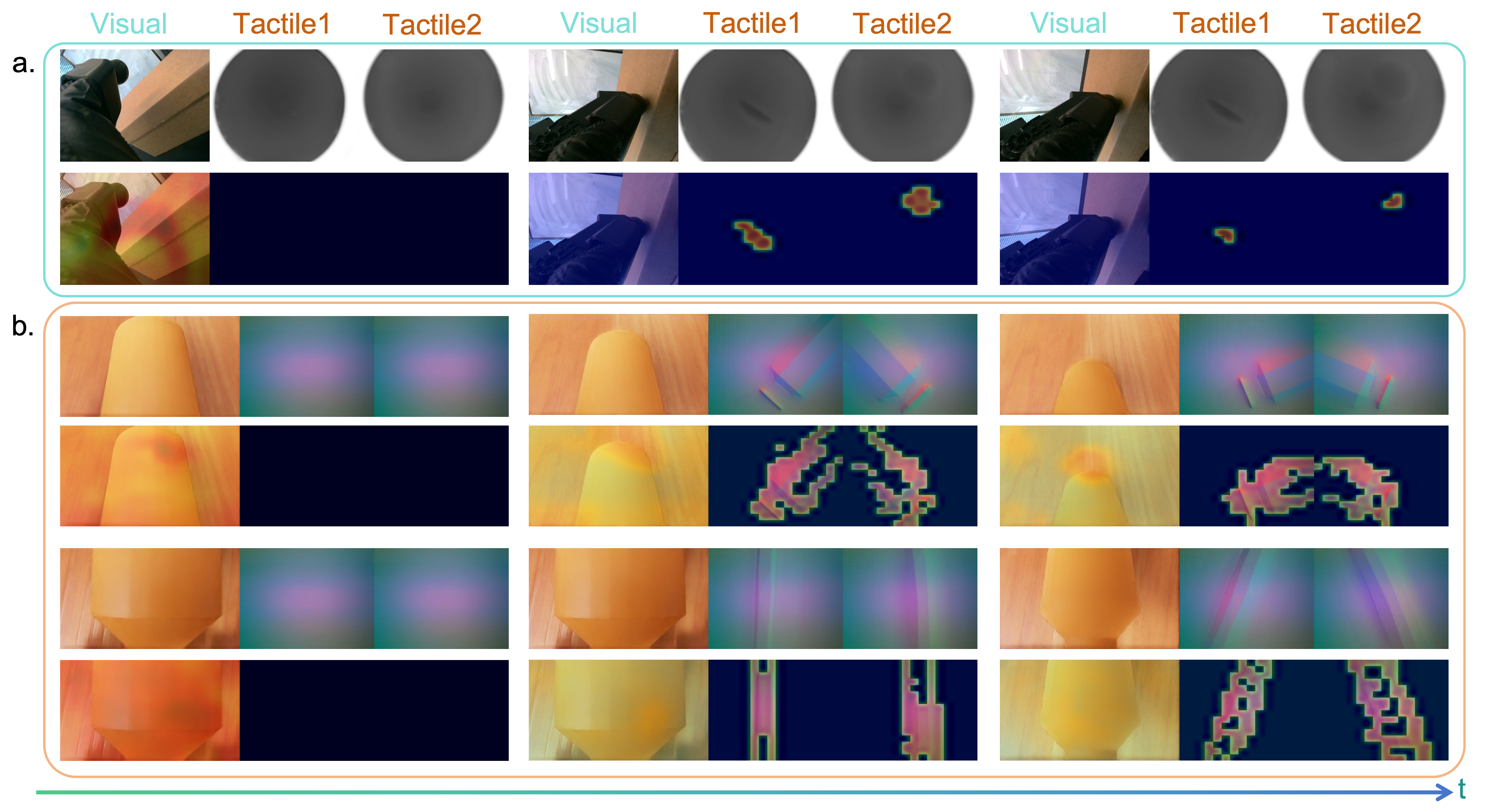}
\caption{Attention heatmap visualization. Panel (a) presents one representative real-world rollout, while panel (b) presents two representative simulated rollouts. Each rollout is arranged in two rows: the top row shows the raw observation frames, and the bottom row shows the corresponding attention heatmaps overlaid on the observations. The heatmaps visualize the modality-specific cross-attention weights after contact-aware attention scaling, with brighter regions indicating tokens with larger scaled attention contributions.}

\label{fig:attn_heatmap}
\end{figure*}

\subsection{Qualitative Analysis of CAAT Attention}
\label{sec:qualitative}

To understand how CAAT modulates multimodal attention across interaction phases, we visualize the Transformer readout attention weights before and during contact. Figure~\ref{fig:attn_heatmap} shows the attention distribution over visual and tactile tokens during a representative rollout. Before contact, the policy attends primarily to visual tokens, leveraging global appearance and spatial information to guide the approach. Once contact occurs, attention shifts toward tactile tokens, allowing the policy to exploit contact-induced deformations and local interaction cues for subsequent manipulation. This phase-dependent shift is consistent with the intended behavior of Contact-Aware Attention Scaling, which directs the policy's attention toward tactile cues during contact.

\subsection{Ablation Studies}
\label{sec:ablation study}

To assess the contributions of CAAT's two main components, we separately analyze Dynamic Tactile Masking and Contact-Aware Attention Scaling in both simulation and real-world settings.

We first evaluate Dynamic Tactile Masking with direct concatenation, learnable gating, and CAAT. As shown in \Cref{tab:mask_ablation}, masking consistently improves all three fusion strategies. For direct concatenation, masking increases the average success rate by $12.8$ and $11.1$ percentage points in simulation and the real world, respectively. Learnable gating exhibits similar gains of $10.0$ and $13.3$ percentage points. The largest improvements are observed for CAAT, with gains of $21.5$ percentage points in simulation and $16.7$ percentage points in the real world. These consistent improvements demonstrate that Dynamic Tactile Masking is effective across different fusion architectures rather than being specific to CAAT. By suppressing static background regions in tactile observations, it enables the policy to focus on informative contact-induced changes.

We further evaluate Contact-Aware Attention Scaling by comparing the full CAAT framework with the fixed-fusion baseline used in the main simulation and real-world comparisons. This baseline retains Dynamic Tactile Masking but removes contact-aware scaling. Removing attention scaling substantially degrades performance in both settings, with a more pronounced reduction in the real-world experiments. This result suggests that dynamically adjusting the relative importance of visual and tactile information is particularly important in the presence of real-world contact variations and sensory uncertainty. Unlike fixed fusion, Contact-Aware Attention Scaling increases the contribution of tactile observations during contact-critical phases while preserving visual guidance during non-contact phases.

Together, these results demonstrate that Dynamic Tactile Masking and Contact-Aware Attention Scaling provide complementary benefits. Dynamic Tactile Masking determines \emph{where} informative tactile changes occur, whereas Contact-Aware Attention Scaling determines \emph{when} tactile information should receive greater emphasis.








\begin{table}[t]
\centering

\begin{tabular}{lccc}
\toprule
\textbf{Fusion Strategy}
& \textbf{Masking}
& \textbf{Sim. Avg.}
& \textbf{Real Avg.} \\
\midrule

\multirow{2}{*}{Direct Concat}
& \xmark & 38.8\% & 21.7\% \\
& \cmark & 51.6\% & 32.8\% \\

\addlinespace
\multirow{2}{*}{Learnable Gating}
& \xmark & 49.6\% & 25.6\% \\
& \cmark & 59.6\% & 38.9\% \\

\addlinespace
\multirow{2}{*}{\textbf{CAAT}}
& \xmark & 48.1\% & 43.3\% \\
& \cmark & \textbf{69.6\%} & \textbf{60.0\%} \\

\bottomrule
\end{tabular}
\caption{Ablation of Dynamic Tactile Masking across visuo-tactile fusion strategies. Masking consistently improves performance by suppressing static tactile background regions and emphasizing contact-relevant changes.}
\label{tab:mask_ablation}
\end{table}

\section{Conclusions}
We presented CAAT, a lightweight framework for data-efficient, contact-rich visuo-tactile manipulation. CAAT encodes the phase-dependent roles of visual and tactile sensing into Transformer-based policies through Contact-Aware Attention Scaling, emphasizing vision before contact and tactile feedback during contact. It further employs Dynamic Tactile Masking to suppress static non-contact backgrounds in tactile observations while preserving contact-induced changes. Together, these two plug-and-play components provide a simple structural prior that enables policies to use tactile information when and where it is most informative.

Experiments in both simulation and real-world settings demonstrate the effectiveness of CAAT. On the UniVTAC benchmark, CAAT improves the average success rate over direct-concatenation and learnable-gating baselines by $18.0$ and $10.0$ percentage points, respectively. Across real-world contact-rich manipulation tasks, CAAT consistently improves ACT, Diffusion Policy, and $\pi_0$, achieving an overall average success rate of $60.0\%$ and outperforming the strongest baseline by $21.1$ percentage points. These results demonstrate that incorporating contact-aware priors can improve data-efficient visuo-tactile policy learning across diverse policy backbones.

\clearpage

\section*{Appendix}

\appendix
\renewcommand{\thesection}{Appendix \Alph{section}}

\section{Hardware Design}
\label{sec:appendix_hardware}

\begin{figure}[h]
    \centering
    \includegraphics[width=\linewidth]{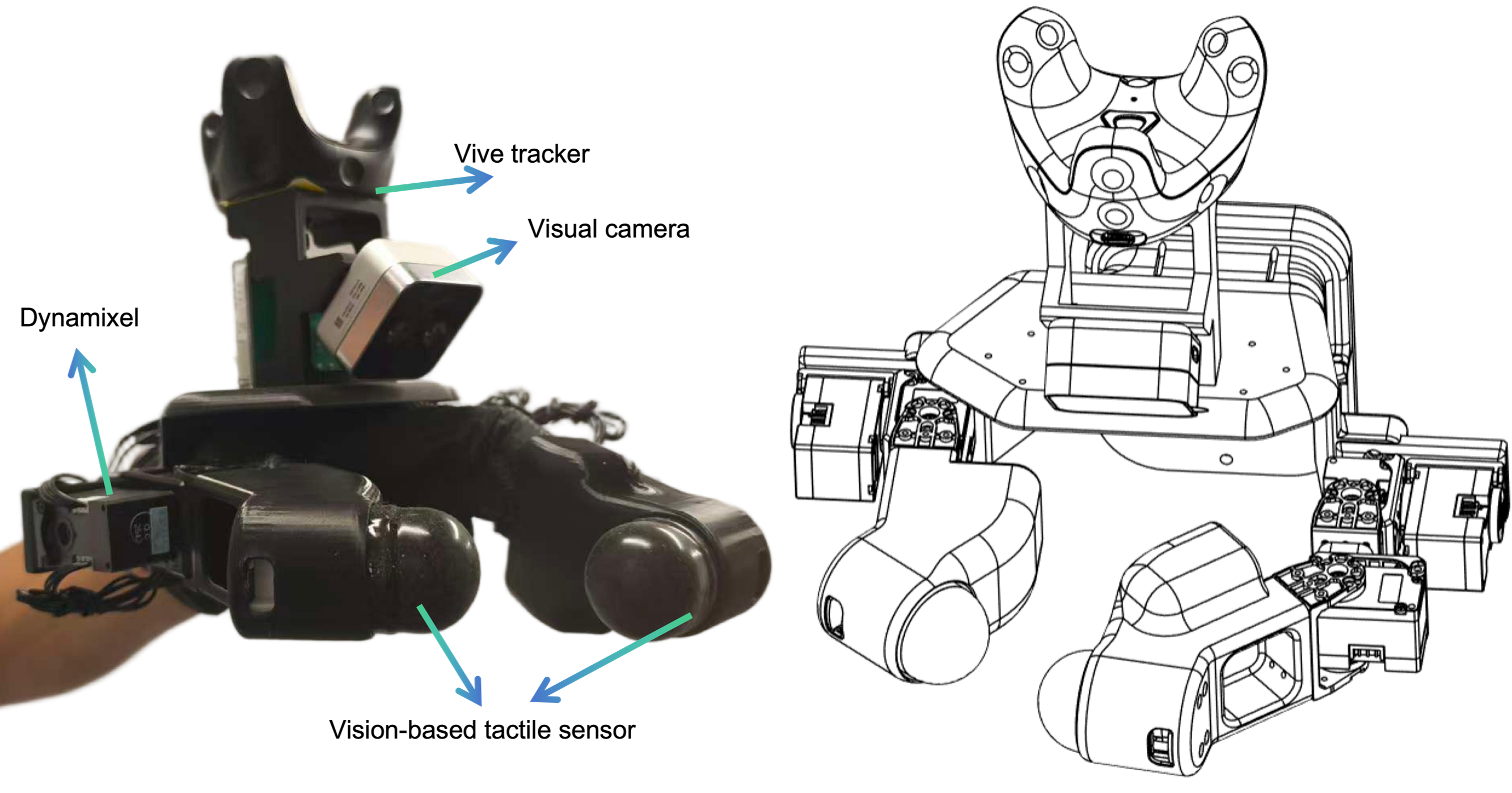}
    \caption{Hardware design of the tactile gripper.
    The dexterous two-finger tactile device is used for both human-operated demonstration collection and robot policy deployment. Each fingertip is equipped with a PPTac tactile sensor, and five Dynamixel motors control the opening/closing and lateral motion of the fingers. During data collection, a Vive Tracker mounted on the gripper records its pose and motion. During deployment, the tracker is removed, and the robot executes the policy based on the end-effector pose.}
    \label{fig:hardware_design}
\end{figure}

Inspired by the UMI data-collection paradigm \cite{chi2024universal}, as well as the structure and pinch-grasp capability of the human hand, we design and manufacture a dexterous two-finger tactile gripper for real-world demonstration collection and policy deployment, as shown in \Cref{fig:hardware_design}. The device can be directly operated by a human demonstrator, enabling intuitive, natural, and repeatable collection of contact-rich manipulation demonstrations. Each fingertip is equipped with a PPTac tactile sensor~\citep{lin2025pp}, which captures high-resolution images of the contact surface during interaction.

The gripper adopts a human-inspired two-finger configuration and is actuated by five Dynamixel XL330 motors. Precise control of the finger opening/closing and lateral motion enables the gripper to manipulate objects of different shapes and sizes. During demonstration collection, a Vive Tracker mounted on the gripper records its pose and motion in real time.

During policy deployment, the Vive Tracker is removed, and the robot executes the predicted actions based on the end-effector pose. Since demonstration collection and policy deployment share the isomorphic gripper hardware and tactile-sensing configuration, the proposed setup reduces the embodiment gap between human demonstrations and robot execution.

\section{Tactile Encoder Details}
\label{sec:appendix_encoder}

The tactile encoder $E_\tau$ used in CAAT is a Vision Transformer (ViT) pretrained on tactile images. We summarize its architecture and pretraining procedure below.

\textbf{Architecture.}
Given a $256 \times 256$ tactile image, the ViT-based encoder produces a $32 \times 32$ grid of spatial features, corresponding to $1{,}024$ tactile tokens. Because policy learning does not require such fine-grained spatial resolution, we apply $2 \times 2$ average pooling to reduce the feature grid to $16 \times 16$. The resulting $N=256$ tactile tokens are subsequently processed by Dynamic Tactile Masking and passed to the policy backbone.

\textbf{Input.} To accommodate different tactile sensors, all input images are resized to a unified resolution of $H_\tau \times W_\tau = 256 \times 256$ pixels. The pretraining corpus contains several hundred thousand images from publicly available datasets, covering diverse manipulation environments and sensor types, including GelSight~\citep{yuan2017gelsight}, DIGIT~\citep{lambeta2020digit}, and 9DTact~\citep{lin20239dtact}.

\textbf{Pretraining.}
We pretrain the tactile encoder using a ViT-VQGAN framework, which learns a discrete tactile codebook through vector quantization and is optimized with reconstruction, perceptual, and adversarial objectives. Training on the large-scale, multi-sensor corpus enables the encoder to learn tactile representations that generalize across tasks and embodiments. After pretraining, the encoder is frozen and requires no task-specific adaptation. For policy learning and Dynamic Tactile Masking, we discard the quantizer and decoder and use the continuous spatial token features produced by the frozen ViT encoder before quantization.

\section{Implementation and Training Details}
\label{sec:appendix_impl}

This appendix provides additional details on the experimental setup, hyperparameters, and training procedure.

\subsection{Policy Backbones}
We evaluate CAAT on three representative Transformer-based imitation learning backbones:

\begin{itemize}
    \item \textbf{ACT} ~\citep{zhao2023learning}: Transformer with 4 encoder layers and 7 decoder layers, hidden dimension 512. The action chunk size is set to 50.
    \item \textbf{Diffusion Policy} ~\citep{chi2025diffusion}: 100 denoising steps with an 8-layer denoising network. The prediction horizon is 10.
    \item \textbf{$\pi_0$} ~\citep{black2024pi_0}: We use the pretrained $\pi_0$ checkpoint and fine-tune with CAAT components.
\end{itemize}

\subsection{Implementation of the Learnable Baseline and CAAT Variant}
\label{sec:learnable_variants}

\paragraph{Learnable-gating baseline.}
The learnable-gating baseline applies a single trainable coefficient to
the visual and tactile tokens before joint self-attention. After the
visual and tactile observations are processed by their respective
encoders and Dynamic Tactile Masking is applied to the tactile stream,
the encoded tokens are weighted as
\begin{equation}
    \bar{X}_t^v = \alpha X_t^v,
    \qquad
    \bar{X}_t^\tau = (1-\alpha) \widetilde{X}_t^\tau,
\end{equation}
where $\alpha=\sigma(a)$, $a$ is an unconstrained learnable scalar, and
$\sigma(\cdot)$ denotes the sigmoid function. The parameter $a$ is
optimized jointly with the policy using the native training objective
of the corresponding backbone. The weighted visual and tactile tokens
are then concatenated with the remaining input tokens and passed through
the policy Transformer for joint self-attention.

The same learned coefficient is shared across all time steps and is not
conditioned on the estimated contact state. Therefore, although the
baseline can learn the overall relative importance of the two
modalities, it does not explicitly adjust their weighting between the
non-contact and contact phases. Moreover, modality weighting is
performed before joint cross-modal processing, whereas CAAT applies
contact-aware scaling to the modality-specific action-query readouts
after joint self-attention.

\paragraph{Learnable CAAT variant.}
Ours (Learnable) preserves the proposed CAAT architecture and differs
from the numerical variant only in the parameterization of the
contact-aware scaling factors. Specifically, the visual and tactile
tokens first undergo joint self-attention. The action query then attends
separately to the contextualized visual, tactile, and proprioceptive
token subsets, producing the modality-specific readouts $r_t^v$,
$r_t^\tau$, and $r_t^\Upsilon$, respectively.

Instead of using predefined numerical scaling factors, we learn a
state-dependent visual scaling coefficient:
\begin{equation}
    \gamma_v^{\mathrm{L}}(z_t)
    =
    \sigma\left(g_{z_t}\right),
\end{equation}
where $z_t \in \{0,1\}$ denotes the estimated contact state,
$g_{z_t}$ is the learnable parameter associated with state $z_t$, and
$\sigma(\cdot)$ denotes the sigmoid function. The corresponding tactile
scaling coefficient is defined as its complementary weight:
\begin{equation}
    \gamma_\tau^{\mathrm{L}}(z_t)
    =
    1-\gamma_v^{\mathrm{L}}(z_t).
\end{equation}
The final action-query representation is computed as
\begin{equation}
    \hat{r}_t
    =
    \gamma_v^{\mathrm{L}}(z_t) r_t^v
    +
    \left[1-\gamma_v^{\mathrm{L}}(z_t)\right] r_t^\tau
    +
    r_t^\Upsilon.
\end{equation}

Unlike the learnable-gating baseline, which uses the same modality
coefficient across all interaction phases and applies it before joint
self-attention, Ours (Learnable) uses contact-state-dependent
coefficients to scale the modality-specific action-query readouts after
joint self-attention. All other components, including Dynamic Tactile
Masking, contact-state estimation, and the action decoder, remain
unchanged. This variant therefore evaluates whether the modality
weights for different contact states can be learned effectively from
demonstrations, in comparison with the explicitly specified numerical
contact-aware prior.

\subsection{Training Hyperparameters}
All models are trained with a batch size of 64. We use the AdamW optimizer 
with a fixed learning rate of $1 \times 10^{-5}$ and weight decay of 
$1 \times 10^{-4}$. $\pi_0$ is fine-tuned using LoRA on two NVIDIA A100 80GB GPUs with a global batch size of 64, while ACT and DP are each trained on a single NVIDIA A100 80GB GPU.

\subsection{CAAT-Specific Settings}
For Contact-Aware Attention Scaling, we set $(\gamma_{v}, \gamma_{\tau})=(0.7, 0.3)$ during the non-contact phase and $(\gamma_{v}, \gamma_{\tau})=(0.3, 0.7)$ during the contact phase, as defined in Eq.(13). The contact estimator is a frozen CNN--MLP binary classifier that takes as input the difference between the current tactile image and a non-contact reference image. It outputs a contact probability in $[0,1]$, with a threshold of $0.5$ used to determine contact. For Dynamic Tactile Masking, we use a cosine-similarity threshold of $\rho=0.8$ to identify tokens corresponding to static tactile background. The tactile image captured at the first timestep of each episode serves as the non-contact reference.

\subsection{Data Collection}

For the real-world tasks, we collect demonstrations via teleoperation using the aforementioned UMI-inspired handheld device. The device is equipped with a two-finger tactile gripper and a Vive Tracker for pose tracking. We collect 150 successful demonstrations for each task. For the simulation tasks, we use the standard UniVTAC dataset~\citep{chen2026univtac}, which also contains 150 demonstrations per task.

\subsection{Evaluation Protocol}

For simulation, we evaluate each task over 100 rollouts across 100 random seeds and report the mean success rate. For real-world evaluation, each task is run 20 times per checkpoint with randomized object poses, and we report the mean success rate across 20 trials.

\section{Complete Data-Efficiency Results}
\label{sec:data_efficiency_details}

\Cref{tab:data_efficiency_results} reports the complete average success
rates across the five UniVTAC tasks under different demonstration
budgets.

\begin{table}[t]
\centering
\small
\begin{tabular}{lcccc}
\toprule
\textbf{Method}
& \textbf{25}
& \textbf{50}
& \textbf{100}
& \textbf{150} \\
\midrule
ACT + DTM
& 21.6 & 25.4 & 22.2 & 51.6 \\
Gated ACT + DTM
& 15.4 & 30.0 & 27.8 & 59.6 \\
\textbf{CAAT (Numerical)}
& \textbf{24.8}
& \textbf{34.0}
& \textbf{38.4}
& \textbf{69.6} \\
\bottomrule
\end{tabular}
\caption{Average success rates (\%) across five UniVTAC tasks under
different numbers of training demonstrations.}
\label{tab:data_efficiency_results}
\end{table}

\bibliography{aaai2027}
\end{document}